\documentclass{bmvc2k}

\usepackage{wrapfig}
\usepackage{enumitem}
\usepackage{xscommand}

\title{Diversifying the High-level Features for better Adversarial Transferability}

\addauthor{Zhiyuan Wang}{vinlee624@gmail.com}{$\ast$1}
\addauthor{Zeliang Zhang}{hust0426@gmail.com}{$\ast$1}
\addauthor{Siyuan Liang}{liangsiyuan@iie.ac.cn}{2}
\addauthor{Xiaosen Wang}{xiaosen@hust.edu.cn}{$\dag$3}
\addinstitution{
School of Computer Science and \\Technology, 
 Huazhong University of \\Science and Technology
}
\addinstitution{
 Institute of Information Engineering \\Chinese Academy of Sciences	
}

\addinstitution{
 Huawei Singularity Security Lab	\\
 \vspace{0.15em}
 \textsuperscript{$\ast$}Equal contribution \\
 \vspace{0.15em}
 \textsuperscript{$\dag$}Corresponding author
}

\runninghead{Wang et al.}{DHF for better Adversarial Transferability}

\def\eg{\emph{e.g}\bmvaOneDot}

\def\etal{\emph{et al}\bmvaOneDot}

\begin{document}

\maketitle

\begin{abstract}
Given the great threat of adversarial attacks against Deep Neural Networks (DNNs), numerous works have been proposed to boost transferability to attack real-world applications. However, existing attacks often utilize advanced gradient calculation or input transformation but ignore the white-box model. Inspired by the fact that DNNs are over-parameterized for superior performance, we propose diversifying the high-level features (\name) for more transferable adversarial examples. In particular, \name perturbs the high-level features by randomly transforming the high-level features and mixing them with the feature of benign samples when calculating the gradient at each iteration. Due to the redundancy of parameters, such transformation does not affect the classification performance but helps identify the invariant features across different models, leading to much better transferability. Empirical evaluations on ImageNet dataset show that \name could effectively improve the transferability of existing momentum-based attacks. Incorporated into the input transformation-based attacks, \name generates more transferable adversarial examples and outperforms the baselines with a clear margin when attacking several defense models, showing its generalization to various attacks and high effectiveness for boosting transferability. Code is available at \url{https://github.com/Trustworthy-AI-Group/DHF}.
\end{abstract}


\section{Introduction}
\label{sec:intro}
Recent studies have shown that Deep Neural Networks (DNNs) are vulnerable to adversarial examples~\cite{szegedy2014intriguing}, \ie, the inputs with indistinguishable perturbation can mislead the DNNs. Such vulnerability brings a significant threat to the security of widely deployed DNNs in the physical world, \eg, image classification~\cite{krizhevsky2012imagenet,szegedy2016rethinking,Karen2015vgg,he2016deep,huang2017densely}, face recognition~\cite{tang2004video,wen2016discriminative,qiu2021end2end,yang2021larnet}, autopilot ~\cite{chabot2017deep,codevilla2018offline, gao2018object, janai2020computer}, \etc. However, existing adversarial attacks exhibit excellent attack performance when the attacker can access the full knowledge of the target model, but show poor transferability across different models, making it inefficient in the real world.

To improve the transferability of adversarial examples, many methods have been proposed, such as input transformations~\cite{xie2018mitigating}, ensemble-model attacks~\cite{liu2016delving}, and model-specific methods~\cite{wu2020skip}. Among them, the model-specific methods, which modify or utilize the inner structure of DNNs, are one of the most effective methods to improve the transferability of adversarial examples and are compatible with the other two types of methods. For instance, Wu~\etal~\cite{wu2020skip} employ the structure of skip connection in ResNet-like DNNs to generate highly transferable adversarial examples. Li~\etal~\cite{li2020learning} propose the ghost network, which virtually ensembles a vast set of diverse models by randomly perturbing the existing model to boost adversarial transferability.  However, both of them lack the full utilization of model structure and image features, remaining potential for further improvements.

Many researchers have pointed out that DNNs are over-parameterization~\cite{banner2018scalable, chmiel2020neural, courbariaux2014training, dong2020hawq, cai2020zeroq}. The increasing number of layers in DNNs brings better feature extraction and classification performance, but correspondingly induces a big redundancy of parameters, especially for the deeper layers. As shown in Fig. \ref{figs:comb_heatmap}, when we randomly mask the high-level features of DNNs, the classification accuracy maintains at a high level. Thus, it naturally inspires us to exploit the over-parameterization of high-level features for better adversarial transferability.  Unlike Wu~\etal~\cite{wu2020skip} and Li~\etal~\cite{li2020learning}, we focus on perturbing only the high-level features and propose a unified perturbing operation in any architectures, not restricted to dropout or scale operation and ResNet-like DNNs. We summarize our contributions as follows:

\begin{itemize}[leftmargin=*,noitemsep,topsep=2pt]
\item To the best of our knowledge, it is the first work that establishes a relationship between over-parameterization and adversarial transferability.
\item We propose a novel approach called Diversifying the High-level Features (\name) that linearly transforms the high-level features and mixes up them with that of benign samples. Such transformation helps identify the invariant features across different models.
    \item Extensive experiments on ImageNet dataset demonstrate that \name can achieve better adversarial transferability than the existing approaches and is general to other attacks.
\end{itemize}


\section{Related Work}

\textbf{Adversarial Attack:}
After Szegedy~\etal~\cite{szegedy2014intriguing} identified the vulnerability of DNNs against adversarial examples, various adversarial attacks have been proposed, \eg, gradient-based attack~\cite{goodfellow2014explaining,kurakin2018adversarial, madry2018towards}, transfer-based attack~\cite{dong2018boosting, xie2019improving,wei2019transferable,long2022frequency}, score-based attack~\cite{ilyas2018black,li2019nattack,chen2017zoo}, decision-based attack~\cite{brendel2017decision, li2020qeba,wang2021triangle}, generation-based attack~\cite{xiao2018generating,wang2019gan}, \etc. Among these, transfer-based attacks do not access the target model, making it popular to attack the deep models in the real world. To improve adversarial transferability, various momentum-based attacks have been proposed, such as MI-FGSM~\cite{dong2018boosting}, NI-FGSM~\cite{Lin2020Nesterov}, VMI-FGSM~\cite{wang2021enhancing}, EMI-FGSM~\cite{wang2021boosting}, \etc. Researchers also propose several input transformation methods (\eg, DIM~\cite{xie2019improving}, TIM~\cite{dong2019evading}, SIM~\cite{Lin2020Nesterov}, Admix~\cite{wang2021admix}, SIA~\cite{wang2023structure}, STM~\cite{ge2023improving}, BSR~\cite{wang2023rethinking}, \etc), which transform the input image before gradient calculation for better transferability.  On the contrary, few works focus on the white-box model itself to craft more transferable adversarial examples. TAP~\cite{zhou2018transferable} maximizes the difference of the feature maps for all layers between the benign sample and adversarial example to enhance the adversarial transferability, while  ILA~\cite{huang2019enhancing} finetunes an adversarial example by enlarging the similarity of the feature difference at a given layer.   Ghost network~\cite{li2020learning} densely adds a dropout layer~\cite{srivastava2014dropout} after each convolutional layer to create a huge set of diverse models for better transferability. Wu~\etal~\cite{wu2020skip} argues that adopting more gradient from the skip connections could boost the transferability of adversarial examples across various ResNet-based models~\cite{he2016deep}. In this work, we try to diversify the high-level feature to craft more transferable adversarial examples, which are general to any DNNs.

\textbf{Adversarial Defense:}
To mitigate the threat of adversarial attacks, various defenses have been proposed, such as adversarial training~\cite{madry2018towards,Tramr2018EnsembleAT,wang2021multi}, input preprocessing~\cite{xie2018mitigating,Naseer_2020_CVPR}, feature denoising~\cite{liao2018defense,xie2019feature,yang2022robust}, certified defense~\cite{raghunathan2018certified,gowal2019scalable,cohen2019certified}, \etc.  JPEG~\cite{guo2017countering} eliminates the adversarial perturbation by applying the JPEG compression to the input image. High-level representation guided denoiser (HGD)~\cite{liao2018defense} trains a denoising autoencoder based on U-Net~\cite{ronneberger2015u} to purify the image. Randomized resizing and padding (R\&P)~\cite{xie2018mitigating} randomly resizes the image and adds padding to mitigate the adversarial effect.  Bit depth reduction (Bit-Red)~\cite{xu2018feature} reduces the number of bits for each pixel to squeeze the perturbation.   FD~\cite{liu2019feature} employs a JPEG-based compression framework to defend against adversarial attacks.  Cohen~\etal~\cite{cohen2019certified} adopt randomized smoothing (RS) to train
a certifiably robust ImageNet classifier.  Neural Representation Purifier (NRP)~\cite{Naseer_2020_CVPR} adopts a self-supervised adversarial training mechanism to eliminate perturbation effectively. 
 
\textbf{Over-parametering for DNNs:} 
Since Krizhevsky~\etal~\cite{krizhevsky2012imagenet} achieved superior performance using Convolutional Neural Networks (CNNs) on ImageNet, DNNs are becoming deeper and wider with millions of parameters~\cite{szegedy2016rethinking,he2016deep,huang2017densely,wu2019wider,nakkiran2021deep}. Over-parameterization, characterized by deeper and wider neuron layers, has been widely recognized to significantly boost performance~\cite{szegedy2016rethinking,he2016deep,huang2017densely,wu2019wider,nakkiran2021deep}. On the other hand, model quantization has been employed to enable the integration of deep neural networks into mobile phones and embedded devices. This technique reduces the memory requirements of over-parameterized models by replacing the floating point weights with low-precision weights~\cite{banner2018scalable, chmiel2020neural, courbariaux2014training, dong2020hawq, cai2020zeroq}. 
The success of model quantization suggests that DNNs contain redundant information, as the decrease in precision does not significantly impact the final results. In this work, we aim to utilize such redundancy to generate more transferable adversarial examples.


\section{Methodology}
In this section, we first provide our motivation and detail our proposed approach called Diversifying the High-level Features (\name). Then we analyze why high-level features are better than low-level features when diversifying the features and summarize the difference between \name and Ghost. 

\subsection{Motivation}

Lin~\etal~\cite{Lin2020Nesterov} analogize the generation of adversarial examples with the training process of models. In this case, the transferability of adversarial examples is equivalent to model generalization. Recent studies have demonstrated that input transformation-based methods, akin to data augmentation in model training, can significantly enhance adversarial transferability~\cite{xie2019improving,dong2019evading,wang2021admix}. On the other hand, Li~\etal~\cite{li2020learning} boosts transferability by densely applying dropout and random scaling operations on the models (Ghost networks) to diversify the inner features. It sheds light on the potential to enhance transferability by diversifying the features. Motivated by the Ghost networks, we further study how to improve transferability through feature diversification.

\begin{figure}
    \begin{minipage}{0.47\textwidth}
        \centering
        \includegraphics[width=\linewidth]{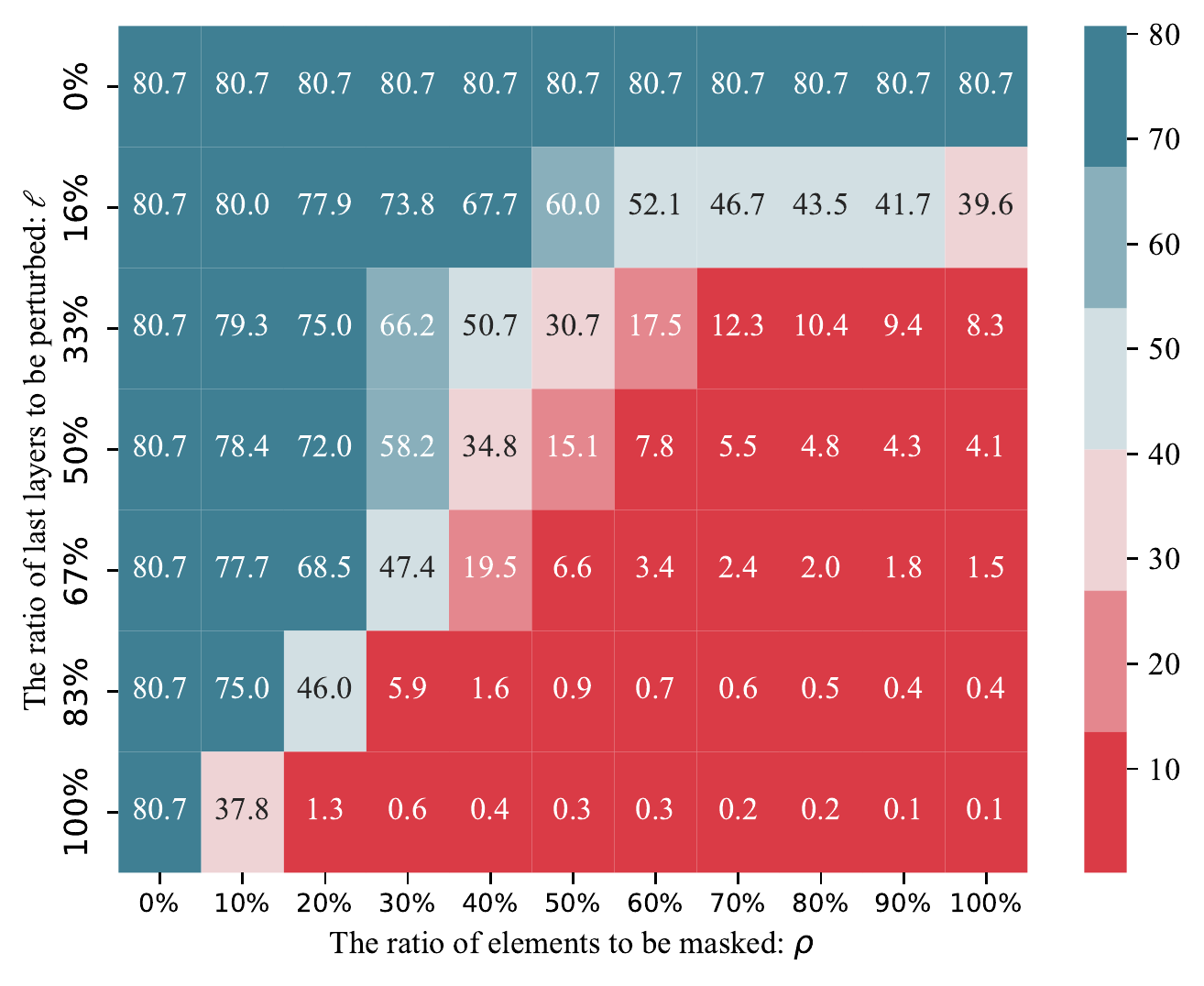}
        \caption{Classification accuracy of Res-101 when randomly masking $\rho$ features of the last $l$ layers among the residual blocks.}
        \label{figs:comb_heatmap}
    \end{minipage}
    \qquad
    \begin{minipage}{0.48\textwidth}
        \centering
        \vspace{1.5em}
        \includegraphics[width=\linewidth]{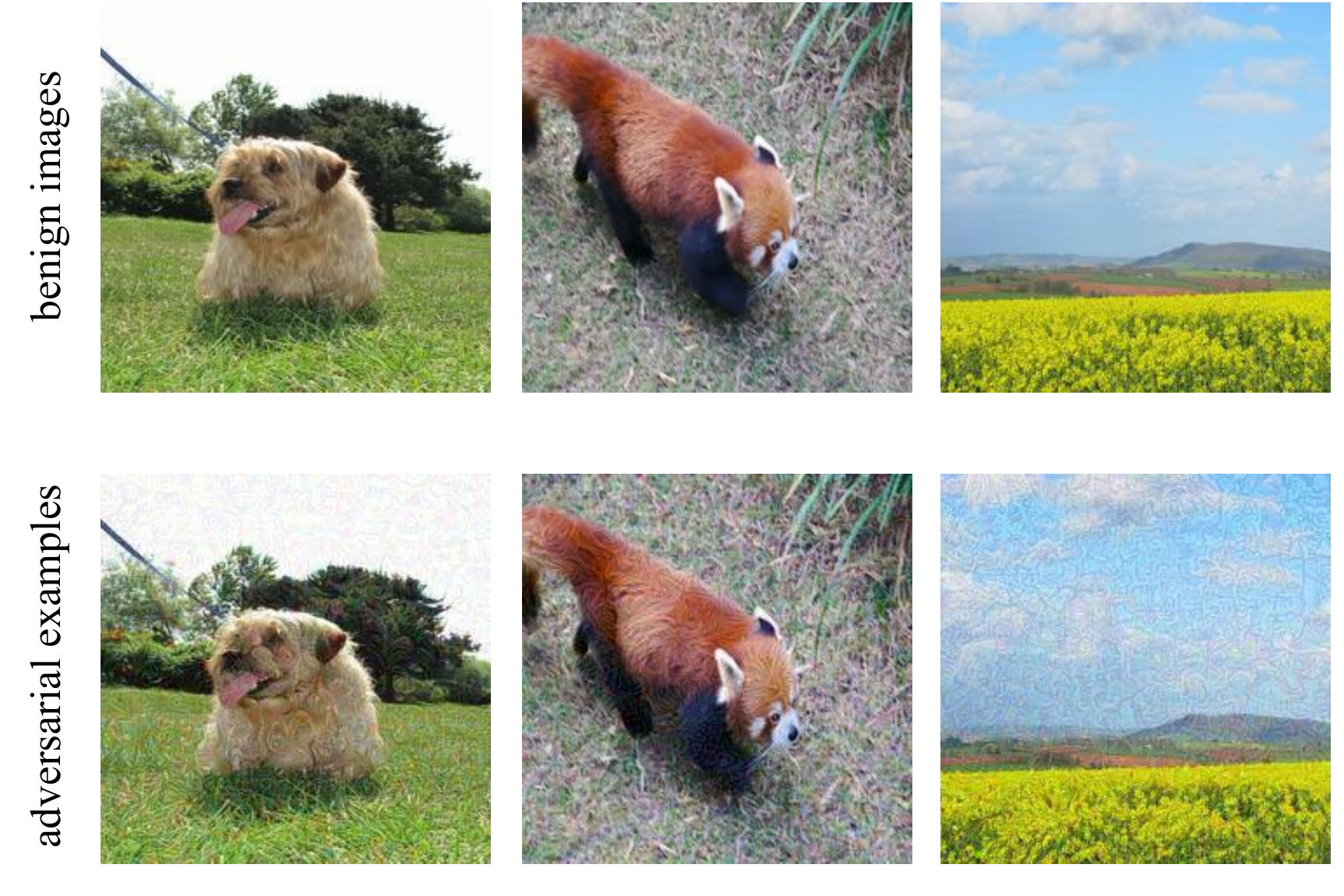}
        \vspace{-0.5em}
        \caption{Visualization of the benign images and their corresponding adversarial examples generated by DHF.}\label{figs:comb_vis_results}
    \end{minipage}
    \label{figs:heatmap_and_vis_results}
\end{figure}

It is widely known that DNNs are over-parameterized for superior performance. Such redundant parameters enable small perturbations among the features to have little impact on the performance, especially in \textit{deeper layers}. To validate this argument, we randomly mask $\rho$ ($0\le \rho \le 1$) elements of features at the last $\mathit{l}$ ($0\le \mathit{l} \le 1$) layers (we only focus on the features of convolutional layers in the residual block) on Res-101~\cite{he2016deep} and report the accuracy in Fig.~\ref{figs:comb_heatmap}. We can observe that randomly masking $10\%$ elements of the last $16.7\%$ layers
does not significantly decay the accuracy. In particular, the accuracy is still $60.0\%$ when we mask half elements of the last $16.7\%$ layers. When we mask more layers or elements at each layer, the accuracy will decrease but can maintain more than $50\%$ most time, highlighting the possibility of diversifying the features using a single model. Moreover, we argue that despite the models' diversity in parameters and structures, they share a similar concentration of features that impacts the performance. Thus, attackers can craft adversarial perturbations on diverse features by perturbing the features, which are excellent for transferring across models. This naturally prompts us to diversify the high-level features to identify the invariant features for better transferability. \ZY{As shown in Fig. \ref{figs:comb_vis_results}, DHF generates high-quality adversarial examples with high transferability. More visualization results can be referred to  Sec. 2 in the supplementary.}

\subsection{Diversifying the High-level Feature}
\label{sub_sec:dhf}

Given a $L$-layer DNN $f$, $y_{\mathit{l}}=f^{(\mathit{l}-1)}(x)$ is the output \wrt the input $x$ at the $\mathit{l}$-th layer ($1\le \mathit{l} \le L$).
To generate highly transferable adversarial examples on $f$, we propose a new approach called Diversifying the High-level Feature (DHF) with two operations detailed as follows:

\textbf{Mixing up the feature.} To diversify the features of adversarial examples without changing the recognition performance, we mix up the features with the features of benign samples:

\begin{equation}
\label{eq1}
    y_{\mathit{l}}^{*} = (1-\eta)\cdot y_{\mathit{l}}^{adv} + \eta \cdot y_{\mathit{l}}, \quad       \eta \sim \mathcal{U}(0, \eta_{max}),
\end{equation}
where $y_{\mathit{l}}=f^{(\mathit{l}-1)}(x)$, $y^{adv}_{\mathit{l}}=f^{(\mathit{l}-1)}(x^{adv})$ and $\mathcal{U}(0, \eta_{max})$ indicates the uniform distribution from 0 to the hyper-parameter $\eta_{max}$. Different from \textit{Admix}~\cite{wang2021admix}, which mixes up the image from other categories, we only mix up the features with that of the benign samples. This diversifies the intermediate layers' features but does not significantly affect the recognition results, making the forward and backward propagation more stable than mixing up some irrelated features. \ZY{Furthermore, as the input samples become adversarial in the attack progress, the gradients of such samples might be unstable since the neighborhood of adversarial examples mainly consists of correctly classified samples. Leveraging benign images helps the optimization process to escape such a region, making the gradient more reliable and stabilizing the optimization.}

\textbf{Randomly adjusting the feature.} As shown in Fig.~\ref{figs:comb_heatmap}, randomly masking the features does not significantly affect the classification accuracy. This validates the redundancy of parameters, which would result in multiple local minima when optimizing the cross-entropy loss for training. Such redundancy would introduce variance into the intermediate features, which differs in the same models trained twice and is more severe across different models, making it difficult for adversarial examples to transfer across models. To eliminate such variance, we randomly replace $\rho$ elements of features at each layer with the mean of features, which stabilizes the propagation and helps boost the transferability. \ZY{Note that replacing the feature with its mean is still differentiable, allowing a more precise gradient w.r.t. the adversarial examples. Compared with adjusting parameters, which will affect the entire feature map, replacing features locally will only change specific elements in the feature map. This distinction makes feature replacement different from parameter adjustment.}

Our proposed \name \ZY{is a general attack applicable
to many existing attacks. It applies the above two operations on the high-level features to other attack methods, \eg MI-FGSM, NI-FGSM, during the forward propagation
of features to prevent the adversarial examples from over-fitting the decision surface of the surrogate model.} These two operations help the attacker eliminate the variance in the features and focus on the invariant features across different models, which stabilizes the gradient calculation and results in highly transferable adversarial examples.

\subsection{Features under Quantitative Approaches}
\label{sec:analysis}
In this section, we analyze why high-level features are better than low-level features when diversifying the features. Without considering the memory footprint of models, the learnable parameters have a specific distribution of float-point numbers. The quantized neural networks reduce the size of the models by limiting the float point numbers to finite values:

\begin{equation}
\label{eq2}
    Q(\theta_{i})=q_j, \theta_{i} \in (r_j, r_{j+1}],
\end{equation}

where $(r_j, r_j+1]$ represents the range of real values depending on the quantization bits $k$, \ie, $j=0,..., 2^k-1$. The quantization process map this range $(r_j, r_j+1]$ to the value $q_j$. 

However, directly using a low-bit precision to quantize a model will decrease the model accuracy. A promising approach to achieve a trade-off between model size and accuracy is to mix different precisions to quantize models. The HAWQ algorithm~\cite{dong2019hawq} utilizes the Hessian eigenvalues as the sensitivity measurement matrix for hierarchical features in the network and adopts mixed-precision quantization for network layers with divergent sensitivities. This motivates us to consider sensitivity when modifying network features. In other words, changing parts of the model at different levels will have different effects on how the model works. To measure network layers, we adopt the average Hessian trace $H$ proposed by the HAWQ-V2 algorithm as the sensitivity matrix. We have the following Lemma:

\begin{lemma}[\cite{dong2020hawq}]
\label{lemma:hawq}
Suppose that the model is twice differentiable and converges to a local minimum when we quantize two layers with the same perturbation, we have:
\begin{small}
\begin{equation}
\label{eq:lemma}
    \mathcal{L}(\theta_1^{*}+\Delta \theta_1^{*}, \theta_2^{*}, ..., \theta_N^{*}) \leq \mathcal{L}(\theta_1^{*}, \theta_2^{*}+\Delta \theta_2^{*}, ..., \theta_N^{*}), 
    \ \mathrm{if} \ 
    \frac{1}{d_1}Tr(\nabla^{2}_{\theta_1}\mathcal{L}(\theta_1^*)) \leq \frac{1}{d_2}Tr(\nabla^{2}_{\theta_2}\mathcal{L}(\theta_2^*)),
\end{equation}
\end{small}
where $d_i$ represents the dimension of the $i$-th parameter $\theta_i$ and $\theta^{*}_i$ represents the converging result of the $i$-th layer in deep networks. The average Hessian trace of the $i$-th layer can be represented by $\frac{1}{d_i}Tr(\nabla^{2}_{\theta_i}\mathcal{L}(\theta_i^*))$.
\end{lemma}

It can be concluded from Lemma~\ref{lemma:hawq} that the average Hessian trace can measure the sensitivity of the layers of the deep model. When the same changes are made, the effect on the loss function is bigger if the average Hessian trace is bigger. As the network layer increases, the average Hessian trace descends, the model becomes less sensitive, and the decision boundary is harder to change, as shown in ~\cite{dong2020hawq}. 

A well-designed quantization scheme does not affect the decision boundary of the model but reduces the computational and memory overhead of the model, indicating that the over-parameterized model has redundancy. Lemma~\ref{lemma:hawq} tells that appropriately changing the point numbers in the weight matrix will not change the loss function. Hence, we have that:

\begin{corollary}
\label{corollary}
Since the over-parameterized network has redundancy, appropriately changing the high-level features will not change the model outputs.
\end{corollary}

With Corollary~\ref{corollary}, we can obtain diverse features by perturbing the high-level features without changing model outputs. While enhancing the diversity of features, we should choose the less sensitive ones so that the deep model's predictions do not change. We can use these diverse features to craft more transferable adversarial examples. Hence, we tend not to change the features of the latter layers.

\subsection{\name \vs Ghost}
\name pays attention to the white-box model by utilizing the redundancy of parameters to perturb the high-level features for better transferability. Li~\etal~\cite{li2020learning} propose to generate a set of Ghost networks by densely applying dropout to each convolutional layer or random scaling on the skip connection for ResNet-based models to enhance the transferability. We summarize the difference between \name and Ghost as follows:

\begin{itemize}[leftmargin=*,noitemsep,topsep=2pt]
    \setlength{\itemsep}{0pt}
    \item \textbf{Motivation.} \name diversifies the features based on the redundancy of parameters to obtain invariant features while Ghost aims to reduce the training cost of ensemble model attack.
    \item \textbf{Strategy.} \name perturbs the high-level features since the parameters of lower layers tend to be less redundant. However, Ghost argues that perturbation on the latter layer cannot provide transferability and densely perturbs the features.
    \item \textbf{Transformation.} \name mixups the feature of current examples and benign samples and randomly replaces the features with their means while Ghost densely adopts a dropout layer for convolutional layer or random scaling on the skip connection.
    \item \textbf{Generalization.} As shown in Sec.~\ref{sec:evaluation}, \name could consistently boost the transferability of various attacks while Ghost sometimes degrades the performance.
\end{itemize}


\section{Experiments}
\label{sec:experiments}
In this section, we conduct empirical evaluations to validate the effectiveness of \name. We specify the experimental setting, evaluate \name using momentum-based as well as input transformation-based attacks, and provide parameter studies.

\subsection{Experimental Setting}

\begin{table*}
\begin{center}
\resizebox{\textwidth}{!}{
\begin{tabular}{clbbbbbbbbb}
\toprule
Attack & Method  & Res-18 & Res-50 & Res-101 & Res-152 & IncRes-v2  & DenseNet-121 & MobileNet & ViT & Swin\\
\midrule
\multirow{6}{*}{MI-FGSM} 
& Org. & 52.0 & 54.1 & 45.7 & 51.3 & 40.5 & 52.6 & 52.3 & 23.0 & 35.1\\
& TAP & 62.8 & 61.6 & 46.5 & 50.8 & 51.9 & 63.9 & 59.0 & 16.7 & 31.7 \\
& ILA & 63.9 & 57.7 & 45.4 & 52.1 & 45.1 & 58.7 & 53.2 & 25.6 & 33.5 \\
& SGM & 64.8 & 73.0 & 47.7 & 53.7 & 51.9 & 70.5 & 63.3 & 29.1 & 45.5 \\
& Ghost & 67.3 & 74.5 & 47.5 & 67.0 & 52.9 & 71.4 & \textbf{65.3} & 28.0 & 45.5 \\
\rowcolor{Gray} \cellcolor{white} & \namea & \setrow{\bfseries} 71.9 & 76.7 & 47.9 & 70.2 & 57.5 & 74.7\clearrow & 62.9 & \setrow{\bfseries} 35.2 & 53.2 \clearrow\\
\midrule
\multirow{6}{*}{NI-FGSM} 
& Org. & 54.8 & 58.9 & 46.8 & 53.0 & 44.0 & 55.9 & 54.2 & 23.9 & 37.6\\
& TAP & 63.4 & 63.9 & 45.7 & 58.2 & 57.6 & 57.3 & 59.7 & 20.1 & 34.5 \\
& ILA & 64.7 & 62.0 & 47.6 & 63.9 & 48.1 & 57.1 & 62.4 & 28.3 & 48.1 \\
& SGM & 65.3 & 75.9 & 47.8 & 55.3 & 50.6 & 73.9 & 66.1 & 30.2 & 48.7 \\
& Ghost & 69.3 & 75.8 & 47.9 & 69.5 & 54.6 & 72.3 & 68.9 & 30.9 & 48.1 \\
\rowcolor{Gray} \cellcolor{white} & \namea & \setrow{\bfseries} 73.0 & 77.3 & 48.5 & 74.8 & 60.3 & 77.1 & 71.5 & 35.8 & 55.9 \clearrow  \\
\bottomrule
\end{tabular}
}
\end{center}
\caption{ Average black-box attack success rates (\%) on nine models by two momentum-based attacks. The adversarial examples are generated on Res-101, Res-152 and IncRes-v2, respectively.}
\label{tab:momentum}
\end{table*}




\textbf{Dataset.} We adopt 1,000 images~\cite{wang2021enhancing} in 1,000 categories, which are randomly sampled from ILSVRC 2012 validation set~\cite{krizhevsky2012imagenet}. All the chosen models can correctly classify the images.

\textbf{Models.} We first evaluate the attack performance on nine models of two popular architectures, namely Convolutional Network Works (CNNs), \ie,
ResNet-18 (Res-18)~\cite{he2016deep}, ResNet-50 (Res-50)~\cite{he2016deep}, ResNet-101 (Res-101)~\cite{he2016deep}, ResNet-152 (Res-152)~\cite{he2016deep}, Inception-ResNet-v2 (IncRes-v2)~\cite{szegedy2017inception}, DenseNet-121~\cite{huang2017densely}, MobileNet~\cite{howard2017mobilenets}, and Transformers, \ie, Vision Transformer (ViT)~\cite{dosovitskiy2020image} and Swin~\cite{liu2021swin}. We also consider several models with defense mechanisms, including the top-3 submissions in the NIPS 2017 defense competition: HGD~\cite{liao2018defense}, R\&P~\cite{xie2018mitigating}, NIPS-r3\footnote{\url{https://github.com/anlthms/nips-2017/tree/master/mmd}}, input preprocessing: Bit-Red~\cite{xu2018feature}, FD~\cite{liu2019feature}, JPEG~\cite{guo2017countering}, a certified defense RS~\cite{cohen2019certified}, and an adversarial perturbation denoiser NRP~\cite{Naseer_2020_CVPR}.

\textbf{Evaluation Settings.} \ZY{To validate the effectiveness of \name, we consider two momentum-based attacks, \ie, MI-FGSM~\cite{dong2018boosting} and NI-FGSM~\cite{Lin2020Nesterov}. We also adopt two input transformations, namely DIM~\cite{xie2019improving} and TIM~\cite{dong2019evading}. DIM adopts the transformation probability  $0.5$, and TIM utilizes the Gaussian kernel with size $7\times7$. Our \name perturbs the last $\frac{5}{6}$s layers with the upper bound of mixup weight $\eta_{max}=0.2$ and the portion of perturbed elements $\rho=10\%$. We set the the perturbation budget $\epsilon = 16$, which
is a typical black-box setting \cite{zhou2018transferable, huang2019enhancing, wu2020skip, li2020learning, dong2018boosting}, and the number of iteration $T = 10$, step size $\alpha = 1.6$ and decay factor $\mu=1.0$.} 

\subsection{Numeric Results}
\label{sec:evaluation}
We evaluate our \name using two momentum-based attacks, namely MI-FGSM~\cite{dong2018boosting} and NI-FGSM~\cite{Lin2020Nesterov}, and two input transformation-based attacks, \ie, DIM~\cite{xie2019improving} and TIM~\cite{dong2019evading}. We adopt the original model and several architecture-related approaches as our baselines, namely TAP~\cite{zhou2018transferable}, ILA~\cite{huang2019enhancing}, SGM~\cite{wu2020skip}, ghost network~\cite{li2020learning}. Since some baselines (\eg, SGM) are limited to ResNets, we take Res-101, Res-152 and IncRes-v2 as the white-box models but \name is general to any architectures. The attack performance is quantified by the attack success rates, which are the misclassification rates of the corresponding models on the generated adversarial examples. \ZY{In Sec.~1.1 of our supplementary, we provide the results of the comparison with more advanced attacks.}

\textbf{Evaluations on Momentum-based Attacks.} We first evaluate the performance improvement of MI-FGSM and NI-FGSM. 

The average attack success rates on the black-box models are summarized in Table~\ref{tab:momentum}. We can observe that all the approaches can boost the adversarial transferability than generating on the original models, in which our \name achieves the best attack performance on either CNNs or transformers. In particular, \name outperforms the best baseline (Ghost) with a clear margin of $3.4\%$ and $4.1\%$ on average for MI-FGSM and NI-FGSM, respectively. With the same computation cost, such superior attack performance and its generality to CNNs and ViTs support the high effectiveness of \name. 

\begin{table*}
\begin{center}
\resizebox{\textwidth}{!}{
\begin{tabular}{ccbbbbbbbbb}
\toprule
Attack & Method  & Res-18 & Res-50 & Res-101 & Res-152 & IncRes-v2  & DenseNet-121 & MobileNet & ViT & Swin\\
\midrule
\multirow{6}{*}{DIM} 
& Org. & 61.2 & 58.7 & 51.5 & 69.5 & 47.7 & 58.9 & 62.2 & 30.6 & 40.5\\
& TAP & 68.4 & 66.9 & 53.9 & 57.1 & 56.6 & 69.0 & 63.9 & 30.1 & 41.1 \\
& ILA & 72.5 & 63.9 & 52.8 & 60.2 & 59.3 & 69.2 & 65.4 & 38.5 & 46.2 \\
& SGM & 74.8 & 75.9 & 56.6 & 64.3 & 62.1 & 75.6 & 71.9 & 40.6 & 54.9 \\
& Ghost & 78.1 & 80.2 & 61.9 & 70.8 & 70.3 & 85.2 & 80.9 & 47.5 & 59.6 \\
\rowcolor{Gray} \cellcolor{white} & \namea & \setrow{\bfseries} 85.3 & 86.9 & 64.2 & 75.8 & 72.9 & 85.7 & 83.6 & 52.9 & 62.7 \clearrow \\
\midrule
\multirow{6}{*}{TIM} 
& Org. & 58.3 & 60.2 & 49.9 & 57.1 & 50.2 & 60.8 & 59.5 & 27.1 & 43.2\\
& TAP & 65.1 & 68.0 & 50.2 & 58.9 & 60.2 & 63.8 & 65.9 & 28.8 & 38.4 \\
& ILA & 69.4 & 65.7 & 51.6 & 65.2 & 62.6 & 64.9 & 65.9 & 40.3 & 52.3 \\
& SGM & 71.9 & 78.3 & 55.6 & 67.5 & 64.3 & 70.0 & 67.2 & 44.5 & 54.2 \\
& Ghost & 76.2 & 81.9 & 63.5 & 68.2 & 69.3 & 79.1 & 75.7 & 48.3 & 58.7 \\
\rowcolor{Gray} \cellcolor{white}  & \name & \setrow{\bfseries} 82.6 & 88.6 & 67.2 & 74.0 & 74.3 & 86.9 & 80.7 & 53.9 & 61.3 \clearrow  \\
\bottomrule
\end{tabular}
}
\end{center}
\caption{Average black-box attack success rates (\%) on nine models by two input transformation-based attacks. The adversarial examples are generated on Res-101, Res-152 and IncRes-v2, respectively.}
\label{tab:input_transformation}
\end{table*}

\textbf{Evaluations on Input Transformation-based Attacks.} Input transformation-based attack is another mainstream approach to boost adversarial transferability. To further validate the effectiveness of \name, we adopt two input transformation-based methods under the same setting, namely DIM, and TIM. These two approaches are integrated into MI-FGSM and the results are presented in Tab.~\ref{tab:input_transformation}. Either DIM or TIM achieves better transferability than MI-FGSM (in Tab.~\ref{tab:momentum}) on the original model and the baselines can consistently boost the transferability. Among these attacks, \name outperforms the runner-up method Ghost on the  adversarial transferability with a notable gap of $4.0\%$ and $5.4\%$ on DIM and TIM, respectively. Such superior performance further supports the high effectiveness of \name and validates our hypothesis that utilizing the redundance of parameters to perturb the high-level features can significantly improve the transferability.

\textbf{Evaluations on Defense Models.} Recently, various defense methods have been proposed to mitigate the threat of adversarial examples. To fully evaluate the effectiveness of \name, we further consider several models with defense mechanisms, including HGD~\cite{liao2018defense}, R\&P~\cite{xie2018mitigating}, NIPS-r3~\footnote{\url{https://github.com/anlthms/nips-2017/tree/master/mmd}}, Bit-Red~\cite{xu2018feature}, FD~\cite{liu2019feature}, JPEG~\cite{guo2017countering},  RS~\cite{cohen2019certified}, and NRP~\cite{Naseer_2020_CVPR}. We generate the adversarial examples using DIM on Res-101, Res-152, and IncRes-v2 models, respectively and test them on the aforementioned eight defense models. We select Res-18 model as the target model for the input processing-based defense, \ie, Bit-Red, FD, JPEG, and NRP. The other four defense approaches adopt the official models provided in the corresponding papers. The average success attack rates are presented in Tab. \ref{tab:defense}. Surprisingly, Ghost, which achieves the best performance among the baselines in the above evaluations, achieves the poorest performance on these defense methods, even poorer than the adversarial examples generated on the original model. By contrast, \name consistently performs better than the baselines and outperforms the runner-up method (SGM) with a clear margin of $5.3\%$ on average, showing its high effectiveness and generality when attacking the black-box models with different defense mechanisms.

\subsection{Parameter Studies}

\name randomly perturbs the high-level features by mixing up the high-level features with clean features and randomly adjusting the mixed features. In this section, we conduct parameter studies to investigate the impact of three hyper-parameters, namely the upper bound of mixup weight $\eta_{max}$, the ratio of elements to be adjusted $\rho$ and the number of layers to be changed. We generate the adversarial examples on Res-101, Res-152, and IncRes-v2, and report the average attack performance on the other eight models.

\begin{table*}[t]
\begin{center}
{\small
\begin{tabular}{cbbbbbbbbb}
\toprule
Method & HGD & R\&P & NIPS-r3 & Bit-Red & FD & JPEG  & RS & NRP   & Average\\
\midrule
Org. & 52.9 & 45.5 & 50.9 & 34.0  & 48.1 & 56.2    & 20.7 & 36.0   & 43.1\\
TAP & 54.1 & 47.2 & 51.2 & 45.1 & 50.6 & 62.4  & 25.5 & 46.4   & 47.8\\
ILA & 56.3 & 48.0 & 53.7 & 47.6 & 52.3 & 64.8  & 22.9 & 46.5   & 49.0\\
SGM & 54.9 & 48.1 & 54.6 & 47.4 & 52.9 & 70.2  & 23.8 & 48.3   & 50.9\\
Ghost & 35.1 & 29.8 & 30.6 & 31.4 & 38.9 & 49.3   & 19.0 & 33.9   & 33.5\\
\rowcolor{Gray} \name & \setrow{\bfseries} 60.1 & 54.7 &55.6 &52.0  &61.7 & 75.3 & 34.3 & 55.7   & 56.2 \clearrow\\
\bottomrule
\end{tabular}
}
\end{center}
\caption{Average black-box attack success rates (\%) by DIM on various defense models. The adversarial examples are crafted on Res-101, Res-152, and IncRes-v2, respectively.}
\label{tab:defense}
\end{table*}

\begin{figure}
    \centering
    \includegraphics[width=\linewidth]{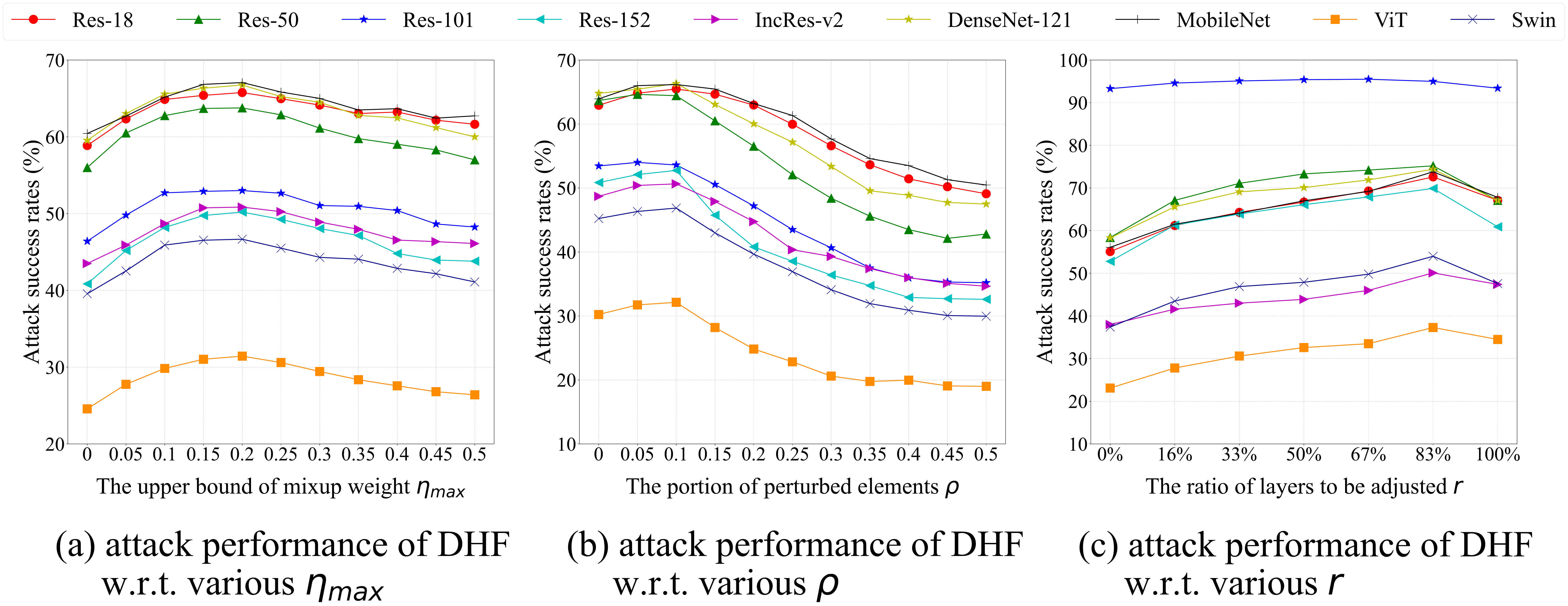}
    \setlength{\abovecaptionskip}{-20pt}
    \caption{Average black-box attack success rates (\%) on nine models by \name using various hyper-parameters. The adversarial examples are generated on Res-101, Res-152 and IncRes-v2, respectively.}
    \label{fig:ablation_params}
\end{figure}

\textbf{On the upper bound of mixup weight $\eta_{max}$}. As shown in Eq.~\eqref{eq1}, $\eta_{max}$ balances the feature of clean sample and adversarial example among the mixed feature. To find a proper value for $\eta_{max}$, we conduct \name attack to perturb the last five-sixths layers with $\eta_{max}$ from $0$ to $0.5$ in a step of $0.05$ using $\rho=0.1$. As shown in Fig.~\ref{fig:ablation_params}~(a), when $\eta_{max}=0$, the mixup operation cannot take effect and \name achieves low transferability. When we increase $\eta_{max}$, the attack performance consistently increases and achieves the peak around $\eta_{max}=0.2$. This further validates that adopting an appropriate portion of clean features can boost adversarial transferability. When $\eta_{max}$ continually increases, the clean features take a more significant portion in the mixed feature, making it hard to calculate the accurate gradient. As a result, the attack performance starts to decay. Hence, we adopt $\eta_{max}=0.2$ in our experiments.

\textbf{On the ratio of elements to be adjusted $\rho$}. $\rho$ reduces the variance among the features to identify the invariant features. To determine the value of $\rho$, we conduct \name attack to perturb the last five-sixths layers with $\rho$ from $0$ to $0.5$ in a step of $0.05$ using $\eta_{max}=0.2$. As shown in Fig.~\ref{fig:ablation_params}~(b), when increasing $\rho$, the overall performance slightly increases when $\rho \le 0.1$, and arrives at the peak when $\rho = 0.1$. Larger $\rho$ will replace more elements with the mean of feature, which decays the classification accuracy as shown in Fig.~\ref{figs:comb_heatmap} and results in the inaccurate gradient. Hence, the attack performance significantly drops and we adopt $\rho=0.1$ in our experiments.

\textbf{On the ratio of layers to be adjusted $r$}. To determine a suitable ratio of layers for adjusting the features, we conduct \name attack to perturb a different number of layers with $\rho=0.1$ and $\eta_{max}=0.2$.

As shown in Fig.~\ref{fig:ablation_params} (c), when \name does not perturb the features (\ie, $r=0\%$), \name cannot take effect and achieves the lowest transferability. When we start to perturb the features (\eg, $r=16\%$), the transferability can be significantly enhanced. The transferability can be consistently improved when \name perturbs more features and achieves the peak when \name perturbs the last $\frac{5}{6}$ layers (\ie, $r=83\%$). Perturbing all the layers results in lower performance than perturbing the last five-sixths layers, which is also consistent with our analysis in Sec.~\ref{sec:analysis}. In our experiments, we perturb the last five-sixths of layers for better performance.

\section{Conclusion}
DNNs are often over-parameterized for good generalization. In this work, we utilize such property to enhance adversarial transferability. Specifically, we find that small perturbations in high-level features  have a negligible impact on overall performance. Motivated by this observation, we propose \name, which perturbs high-level features by randomly transforming and mixing them with benign sample features during gradient calculation. We also provide a theoretical analysis of why high-level features are better than low-level features when diversifying the features. Our extensive evaluations demonstrate that the proposed method achieves significantly better adversarial transferability than existing state-of-the-art attacks.


\bibliography{egbib}










\newpage
\appendix
In this supplementary material, we report the comparison results of DHF and other baselines and the results under different perturbation budgets. Besides, we present some visualization results of benign images and adversarial examples generated by DHF. 
\section{Additional Experiment Results}

\subsection{Comparison with More Baselines}
DHF modifies the feature calculation in forward propagation. We group it into surrogate refinement attacks. Therefore, we compare it with other surrogate refinement attacks, \ie, TAP~\cite{zhou2018transferable}, ILA~\cite{huang2019enhancing}\, SGM~\cite{wu2020skip}, ghost network~\cite{li2020learning} in our paper.

\renewcommand*{\thetable}{\Alph{table}}
\setcounter{table}{0}
\begin{table}[h]
\begin{center}
\resizebox{\linewidth}{!}{
\begin{tabular}{clbbbbbbbbb}
\toprule
&Method  & Res-18 & Res-50 & Res-101 & Res-152 & IncRes-v2 & DenseNet-121 & MobileNet & ViT & Swin\\
\midrule
\multirow{6}{*}{MI-FGSM}
&AA &  58.7& 52.3 & 45.9 & 57.2 & 47.8 & 58.6 & 56.7 & 28.7 &  41.9 \\
&FIA &  67.3& 60.0 & 47.3 & 58.0 & 49.4 & 66.1 & 57.6 & 33.9 & 45.0 \\
&NAA &  68.9& 60.3 & 47.0 & 55.4 & 48.7 & 68.3 & 56.9 & 33.2 & 40.6\\
&LTAP & 68.5 & 64.3 & 46.6 & 59.7 & 50.1 & 68.1 & 58.1 & 33.3 & 47.0\\
&BIA & 61.3 & 58.9 & 44.6 & 59.2 & 48.8 & 62.3 & 55.1 & 30.6 & 44.2\\
\rowcolor{Gray} \cellcolor{white}&\namea & \setrow{\bfseries} 71.9 & 76.7 & 47.9 & 70.2 & 57.5 & 74.7  & 62.9 & 35.2 & 53.2 \clearrow\\
\bottomrule
\end{tabular}
}
\vspace{-0.5em}
\caption{Average black-box attack success rates (\%) on nine models. The adversarial examples are generated on Res-101, Res-152
and IncRes-v2, respectively.}
\label{tab:supp_baselines}
\end{center}
\end{table}

Meanwhile, there are some similar but distinct methods: 1) AA~\cite{inkawhich2019feature}, FIA~\cite{wang2021feature} and NAA~\cite{zhang2022improving}. They are feature disruption attacks, which also adjust the features of adversarial images but focus on the feature distance when optimizing the perturbation; 2) LTAP~\cite{salzmann2021learning} and BIA~\cite{zhang2021beyond}. They have similar mechanism with DHF but they focus on cross-domain transferability (\eg, Cartoon → ImageNet) using pretrained generators, while  DHF focuses on cross-model transferability (\eg, Inc-v3 → ResNet-18).

To help us better understand the mechanism of DHF and illustrate the effectiveness of DHF, we extend to compare DHF with these similar but different methods in Tab.~\ref{tab:supp_baselines}. We observe that DHF surpasses AA, FIA, NAA, LTAP and BIA by $11.3\%$, $7.3\%$, $7.8\%$, $6.0\%$ and $9.5\%$ on average, respectively. The results further validate the superiority of DHF.

\subsection{Results when Perturbation Budget $\epsilon=8$}
The setting of perturbation budget $\epsilon = 16$ is general for transfer-based attacks. Some works also take the perturbation budget $\epsilon = 8$ as an optional setting \cite{li2020learning, xie2019improving}. To fully validate the effectiveness of DHF, we compare DHF with the $2$ SOTA baselines, \ie, SGM, and ghost network when $\epsilon = 8$. The results are summarized in Tab.~\ref{tab:supp_eps_8}. Despite of the reduced perturbation budget, DHF still outperforms the
baselines by a significant margin, showing its high effectiveness.

\renewcommand*{\thetable}{\Alph{table}}
\setcounter{table}{1}
\begin{table}[tb]
\begin{center}
\resizebox{\linewidth}{!}{
\begin{tabular}{clbbbbbbbbb}
\toprule
&Method  & Res-18 & Res-50 & Res-101 & Res-152 & IncRes-v2 & DenseNet-121 & MobileNet & ViT & Swin\\
\midrule

\multirow{3}{*}{
MI-FGSM
($\epsilon=8$)}
& SGM & 39.3 & 34.1 & 26.1 & 28.3 & 26.2 & 35.4 & 41.7 & 14.0 & 23.6 \\
& Ghost & 34.7 & 35.3 & 34.5 & 30.0 & 25.5 & 37.4 & 34.6 & 12.5 & 23.0\\
\rowcolor{Gray} \cellcolor{white}&\namea & \setrow{\bfseries} 41.4 & 44.0 & 43.3 & 38.7 & 30.1 & 44.7 & 42.1 & 19.0 & 29.7 \clearrow\\
\bottomrule
\end{tabular}
}
\vspace{-0.5em}
\caption{Average black-box attack success rates (\%) on nine models. The adversarial examples are generated on Res-101, Res-152
and IncRes-v2, respectively, when $\epsilon=8$.}
\label{tab:supp_eps_8}
\end{center}
\end{table}

\renewcommand*{\thefigure}{\Alph{figure}}
\setcounter{figure}{0}
\begin{figure}[h]
    \centering
    \includegraphics[width=\linewidth]{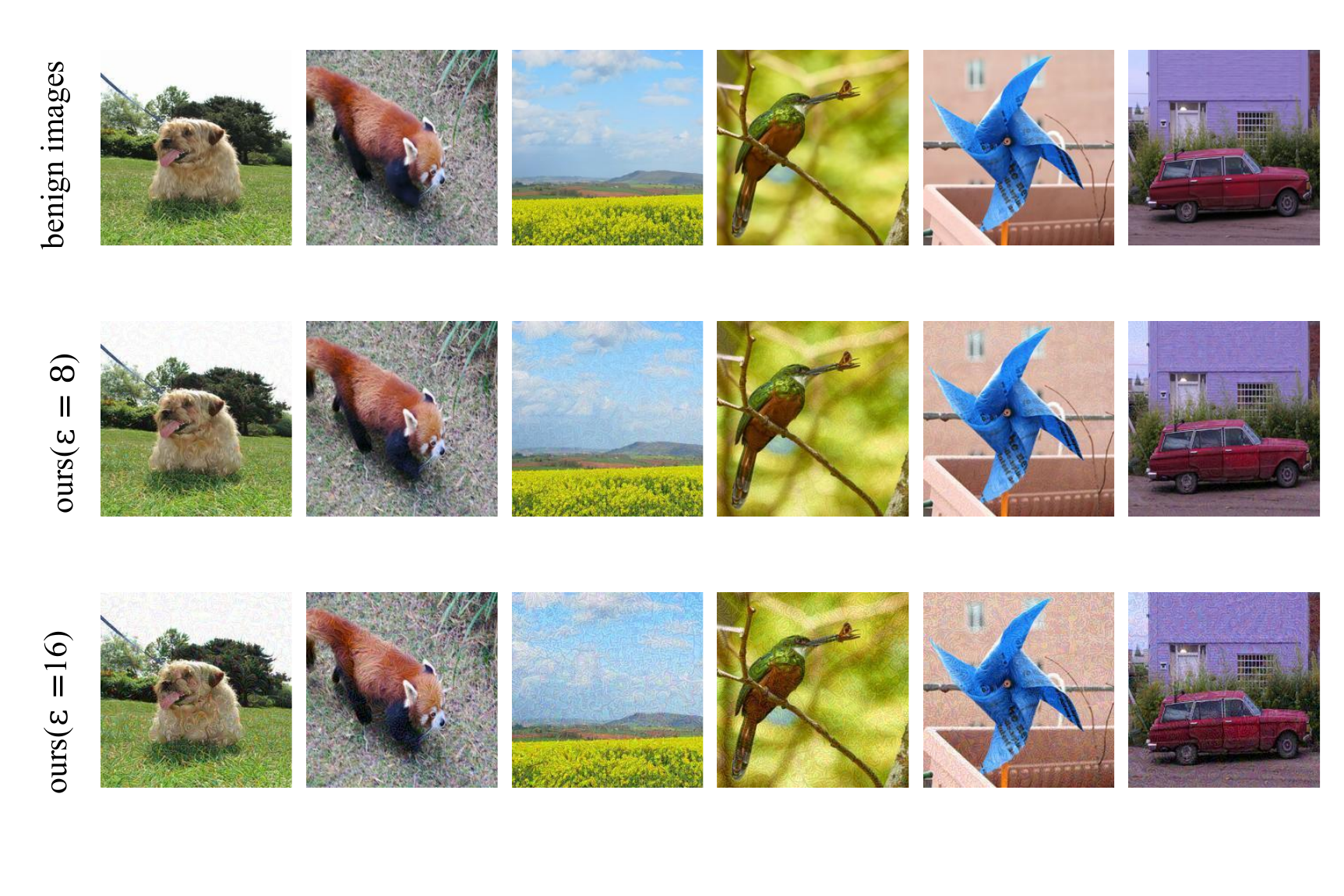}
    \vspace{-3em}
    \caption{Visualization of benign images and the adversarial examples generated by DHF when the perturbation budget $\epsilon=8$ and $\epsilon=16$, respectively. 
    }
    \label{fig:supp_vis}
\end{figure} 

\section{Visualization Reuslts}

In Fig.~\ref{fig:supp_vis}, we present some visualization results of the adversarial examples generated by DHF when $\epsilon = 8$ and $\epsilon=16$, respectively. The adversarial examples exhibit a remarkable visual similarity to the benign images with high adversarial transferability.



\end{document}